\RequirePackage{fix-cm}
\documentclass[twocolumn]{svjour3}       % onecolumn (second 
\smartqed  % flush right qed marks, e.g. at end of proof
\usepackage{graphicx}
\usepackage{soul,color}
\newcommand{\mathcolorbox}[2]{\colorbox{#1}{$\displaystyle #2$}}

\renewcommand{\hl}[0]{}
\renewcommand{\mathcolorbox}[2]{#2}

\usepackage{natbib} % loaded by svjour3 with natbib option
\usepackage[utf8]{inputenc} % allow utf-8 input
\usepackage[draft]{hyperref}       % hyperlinks
\usepackage{url}            % simple URL typesetting
\usepackage{booktabs}       % professional-quality tables
\usepackage{nicefrac}       % compact symbols for 1/2, etc.
\usepackage{microtype}      % microtypography
\usepackage{graphicx} 
\usepackage{subfigure} 
\usepackage{amsmath}
\usepackage{bm}
\usepackage{amsfonts}
\usepackage{amssymb}

\usepackage{hyperref}
\usepackage{algpseudocode,algorithm,algorithmicx}
\algdef{SE}[DOWHILE]{Do}{doWhile}{\algorithmicdo}[1]{\algorithmicwhile\ #1}%
\begin{document}

\title{Gaussian Process Regression for Binned Data\thanks{This work has been supported by the Engineering and Physical Research Council (EPSRC) Project EP/N014162/1. We thank Wil Ward and Fariba Yousefi for their assistance \& suggestions.}}
%\subtitle{}


\author{Michael Thomas Smith \and Neil D Lawrence\footnote{.}\thanks{${}^\star$Work conducted while at the University of Sheffield.} \and Mauricio A \'{A}lvarez}


\institute{Michael Thomas Smith \at
Department of Computer Science\\
University of Sheffield\\
\email{m.t.smith@sheffield.ac.uk}           
           \and
Neil D Lawrence \at
Department of Computer Science\\
University of Sheffield\\
\email{neil@sheffield.ac.uk}  
           \and
Mauricio A \'{A}lvarez L\'{o}pez \at
Department of Computer Science\\
University of Sheffield\\
\email{mauricio.alvarez@sheffield.ac.uk}            
}

\date{Received: date / Accepted: date}

\maketitle

\begin{abstract}
Many datasets are in the form of tables of binned data. Performing regression on these data usually involves either reading off bin heights, ignoring data from neighbouring bins or interpolating between bins thus over or underestimating the true bin integrals.

In this paper we propose an elegant method for performing Gaussian Process (GP) regression given such binned data, allowing one to make probabilistic predictions of the latent function which produced the binned data.

We look at several applications. First, for differentially private regression; second, to make predictions over other integrals; and third when the input regions are irregularly shaped collections of polytopes.

In summary, our method provides an effective way of analysing binned data such that one can use more information from the histogram representation, and thus reconstruct a more useful and precise density for making predictions.

\end{abstract}

\section{Introduction}

Consider the following problem. You want to use a dataset of children's ages and heights to produce a prediction of how tall a child of 38 months will be. The dataset has been aggregated into means over age ranges: e.g. those aged 24 to 36 months have an average height of 90cm, those aged 36 to 48 months, 98cm, etc.

A naive approach would be to simply read off the age range's mean. A slightly more advanced method could interpolate between bin centres. The former method fails to use the data in the neighbouring bins to assist with the prediction, while the latter will produce predictions inconsistent with the dataset's totals. Ideally we would have access to the original dataset, however binning such as this is ubiquitous, sometimes for optimisation (for storage or processing, for example hectad counts in ecology, annual financial reports, traffic counts), sometimes as an attempt to preserve privacy (for example geographical and demographic grouping in the census) and sometimes due to the data collection method itself (camera pixels or fMRI voxels; survey selection, as in section \ref{audience}; or rain-gauge measurements taken each hour). The examples in this paper cover some of these use cases, although many others exist. We also demonstrate how this method can be combined with differential privacy (DP), to provide a simple method for performing DP-regression.

\hl{This problem is a particular example of symbolic data analysis (SDA); in which a latent function or dataset (micro-data) is aggregated in some way to produce a series of symbols (group level summaries). In SDA inference is then conducted at the symbol-level} \citep{beranger2018new}. \hl{It often ignores the underlying likely distributions, often assuming that the data lies uniformly within the histogram bins} \citep{le2017principal} \hl{for example}.

In this paper, we propose a method for performing Gaussian Process (GP) regression given such binned data. Gaussian Processes are a principled probabilistic method for performing regression. Put simply they provide a way of describing how one believes data across input space is correlated, and thus make predictions using previously given training data. We show how one can find the correlation between a cuboid region's integral and a single point. In essence, the histogram we are working with can be considered as a simple form of density estimation. We attempt to extract as much information as possible from the histogram representation, and reconstruct a more useful and precise density.  We will later refer to the analytically derived kernel as the \emph{integral kernel}.

The analytical method described only applies when the boundaries of the integral are independent of one another (i.e. the volume to be integrated over is a cuboid) which often occurs in real datasets, for example in population surveys one might bin people into age ranges and income ranges. However there are many cases where the bins are non-cuboid. An obvious example is that of census tracts, which often follow complicated paths. \citet{kyriakidis2004geostatistical} handles this by approximating the volumes with sets of points, over which the covariance is computed. We briefly re-examine this method (including the problem of point placement and hyperparameter optimisation), extending it to use a set of hyperrectangles to approximate the volumes instead.

The analytical derivation is based on work by \citet{alvarez2009latent} in which dynamical systems are modelled using a Gaussian process. A related article is that of \citet{avzman2005comprising} in which derivatives of a function are observed (rather than the function itself). This was later used to enforce monotinicity in \citet{riihimaki2010gaussian}. A similar constraint was described in \citet{gosling2007nonparametric} who use the observation that a GP and its derivative are jointly Gaussian. In the current paper we operate in the opposite direction, and assume we have observations of a collection of definite integrals of a latent function. \citet{oakley2007uncertainty} follow a similar line of reasoning to this paper and use this to sample the posterior using MCMC. Our method has the advantage that it has a closed form solution, but the use of MCMC allowed them to integrate over the model's hyperparameters and enforce other constraints (such as non-negative latent functions, for example if they are describing a probability density function). Our method operates over higher dimensions and its speed means it can be used as part of an approximation for non-rectangular regions, as in Section \ref{rectangle_approximation_section}. \hl{The work of }\citet{calder2007some}\hl{ is also related, as it focuses on the creation of new kernels though integration. In their case however it was to apply convolution to the latent function.}

Note that, unlike probabilistic integration \citep{o1991bayes}, we are not trying to integrate a function, but rather have been given the integrals of an unknown `latent' function and wish to reconstruct this unknown function. %In sections \ref{shapes} and \ref{audience} we predict integrals from this latent function.

%Quick blurb about Functional Data Analysis - bit wrong?
There is a slight relationship with the combining of basis functions in functional analysis. In \citet[equation 16.3]{ramsay2006functional}, the authors describe how a new kernel is created by summing over the basis functions of two one-dimensional bases, \hl{somewhat like how} we integrate over the (effectively infinite, for a GP) basis functions that lie over the domain being integrated.
%A Gaussian Process regression mean, can be considered as the weighted sum of a series of kernels, placed over each training point. Thus the method we introduce below has some similarity to functional linear models (in the field of functional analysis). Specifically, in the construction of a basis function over the response and covariate. The double integral over the latent function's kernel in our paper (eq?) has similarities to the double sum over the basis functions for the individual variables of temperature and rainfall in (16.3). In the textbook the bases were Fourier, but could easily have been exponentiated quadratics (EQs). An important difference however follows from the method of weighting these bases. In functional analysis one might use a fitting criterion to compute the coefficients B, while in Gaussian process regression, the kernel `weights' are (using the representer theorem) $alpha = K^{-1} \mathbf{y}$. Second, the basic functional analysis result, minimising a fitting criterion leads to a single function, but by using a Gaussian process, we are effectively considering the outputs to be a stochastic process of correlated random variables in which we make predictions using the posterior of these correlations conditioned on observations. This allows us to include the (co)variance of the predictions. We also are able to make use of the regularisation provided by the prior covariance between random variables.

We derive both the analytical and approximate forms of the kernel, then demonstrate the methods on a series of simulated and real datasets and problems.

\section{Analytical Derivation}
\label{analytical_derivation}
To begin we consider the analytical formulation in which we believe that there is a latent function that has been integrated to provide the outputs in the training data. We assume, for now, that we want to make predictions for this latent function. To proceed via Gaussian process regression \citep{williams2006gaussian} \hl{and continuing with our child-height example}, we assume that there is some \emph{latent function}, $f(t)$, that represents the values of \hl{height} as a function of age. The summary measures (average over age ranges) can then be derived by integrating across the latent function to give us the necessary average. \hl{Importantly, if the latent function is drawn from a Gaussian process then we can construct a Gaussian process from which both the latent function and its integral are \emph{jointly} drawn. This allows us to analytically map between the aggregated measure and the observation of interest. }

\hl{To summarise, we assume that a second function, $F(s,t)$, describes the integral between the ages $s$ and $t$ of $f(\cdot)$ and we are given observations, $y(s,t)$, which are noisy samples of $F(s,t)$.} %A note on constructing the training data. In the case above, the values of $y$ are the integrals over the means for each bin; i.e. they are the mean times the width of the bin ranges, for example\\ $y(36,48) = 92 \times 12 = 1104 \; \text{cm months}$.

\hl{A Gaussian process assumption for a function specifies that for a set of random variables, the outputs are jointly distributed as a Gaussian density with a particular mean and covariance matrix. The integration operator effectively adds together an infinite sum of scaled covariances. In summary there will be a Gaussian process with a covariance which describes individually and jointly, the two functions $f(t')$ and $F(s, t)$.} Such a Gaussian process is specified, \emph{a priori}, by its mean function and its covariance function. The mean function is often taken to be zero. It is the covariance function where the main interest lies.

% A Gaussian density has the property that any linear combination of its samples will, in turn, be jointly distributed as Gaussian with the original density. Similarly for a Gaussian process, any \emph{linear operator} (such as integration) applied to the original function will lead to a joint Gaussian process over the result of that linear operator and the original function. In other words there will be a joint Gaussian process between the two functions $f(t')$ and $F(s, t)$. Such a Gaussian process is specified, \emph{a priori}, by its mean function and its covariance function. The mean function is often taken to be zero (although non-zero mean functions are easily incorporated into the framework) but it is the covariance function where the main interest lies.
%
To construct the joint Gaussian process posterior we need expressions for the covariance between values of $f(t)$ and $f(t^\prime)$, values of $F(s,t)$ and $F(s^\prime,t^\prime)$ (i.e. the covariance between two integrals) and the `cross covariance' between the latent function $f(t^\prime)$ and the output of the integral $F(s,t)$. Where $t$, $t^\prime$, $s$ and $s^\prime$ specify input locations.

For the underlying latent function we assume that the covariance between the values of the latent function $f(\cdot)$ is described by the exponentiated quadratic (EQ) form,
$$
k_{ff}(u,u^\prime) = \alpha\; e^{-{{(u-u^\prime)^2} \over {l^2}}},
$$
where \hl{$\sqrt{\alpha}$} is the scale of the output and $l$ is the (currently) one-dimensional length-scale.\footnote{There is a $\sqrt{2}$ difference between our length-scale and that normally defined, this is for convenience in later integrals. Note that other kernels could be substituted, with associated work to integrate the kernel's expression. The supplementary contains a demonstration using the exponential kernel instead.} We are given training points from the integral $F(s,t) = \int_s^t f(u) du$. Reiterating the above, if $f(u)$ is a GP then $F(s,t)$ is also a GP with a covariance we can compute by integrating the covariance of $f(u)$,
$$
\mathcolorbox{yellow}{
k_{FF}((s,t), (s^\prime, t^\prime)) = \; \int_s^t \int_{s^\prime}^{t^\prime} k_{ff}(u,u^\prime)\; \text{d}u^\prime \text{d}u.}
\label{KFFa}
$$
Substituting in our EQ kernel, and integrating,
%note:http://www.progearthplanetsci.org/docs/fomula_template.pdf
% When dividing at a multiplication
%point, the multiplication symbol × should be inserted.
\begin{multline}
 k_{FF}((s,t), (s^\prime, t^\prime)) = \frac{1}{2} \sqrt{\pi } l \alpha  \Biggl[{(s^\prime-s)}\; \text{erf}\left(\frac{s-s^\prime}{l}\right)\\+(s-t^\prime)\;
    \text{erf}\left(\frac{s-t^\prime}{l}\right)+s^\prime\; \text{erf}\left(\frac{s^\prime-t}{l}\right)\\+t\;
    \text{erf}\left(\frac{t-s^\prime}{l}\right)+(t-t^\prime)\; \text{erf}\left(\frac{t^\prime-t}{l}\right)\\+\frac{l}{\sqrt{\pi }}\;
    \biggl(-e^{-\frac{(s-s^\prime)^2}{l^2}}+e^{-\frac{(s-t^\prime)^2}{l^2}}+e^{-\frac{(s^\prime-t)^2}{l^2}}-e^{-\frac{(t-t^\prime)\;
    ^2}{l^2}}\biggr)\Biggr],
\end{multline}
where $\text{erf}(\cdot)$ is the Gauss error function. For ease of interpretation and later manipulation we rewrite this as,
\begin{multline}
k_{FF}((s,t),(s^\prime,t^\prime)) = \alpha \frac{l^2}{2} \\ \;\;\;\;\;\;\;\;\;\;\;\;\;\;\; \times \biggl[ g \left(\frac{t-s^\prime}{l} \right) + g \left(\frac{t^\prime-s}{l} \right) \\- g \left(\frac{t-t^\prime}{l} \right) - g \left(\frac{s-s^\prime}{l} \right) \biggr]
\label{kFF}
\end{multline}
where we defined $g(z) = z\sqrt{\pi}\text{erf}(z)+e^{-z^2}$.

Because we are interested in computing a prediction for the latent function (i.e. the density) that's been integrated, it would be useful to have the cross-covariance between $F$ and $f$. If we assume that the joint distribution of $F$ and $f$ is normal, we can calculate the cross-covariance,
\begin{multline}
\mathcolorbox{yellow}{
k_{Ff}((s,t),(t^\prime)) = \alpha \frac{\sqrt{\pi } l}{2}} \\ \mathcolorbox{yellow}{\times \left(\text{erf}\left(\frac{t-t^\prime}{l}\right)+\text{erf}\left(\frac{t^\prime-s}{l}\right)\right)}.
\label{kFf}
\end{multline}

When using this `integral kernel' in a real GP regression problem we are likely to need to select appropriate hyperparameters. Typically this is done using gradient descent on the negative log marginal-likelihood, $L$, with respect to the hyperparameters. In this case, we need the gradient of $k_{FF}$ wrt $l$ and $\alpha$ (respectively, the lengthscale and variance of the latent EQ function).\footnote{These gradients are then multiplied by $\frac{\partial L}{\partial k_{FF}}$ by the GP framework to give the gradients $\frac{\partial L}{\partial l}$ and $\frac{\partial L}{\partial \alpha}$.} Defining $h(z) = \frac{z\;\sqrt{\pi}}{2}\;\text{erf}(z)\;+\;e^{-z^2}$, we can write the gradient as
\begin{multline}
 \frac{\partial k_{FF}((s,t),(s^\prime,t^\prime))}{\partial l} = \alpha l \\\;\;\;\;\;\;\;\; \times \Big[h \Big(\frac{t-s^\prime}{l}\Big) + h \Big(\frac{t^\prime-s}{l}\Big) \\ - h \Big(\frac{t-t^\prime}{l}\Big) - h \Big(\frac{s-s^\prime}{l}\Big) \Big].
\end{multline}

Similarly we can compute the gradient of the hyperparameters with respect to the cross-covariance ($k_{Ff}$). Defining another support function $d(z) = \frac{z \sqrt{\pi}}{2} \text{erf}(z) - z e^{-z^2}$, we can show that the gradient is 
\begin{multline}
 \frac{\partial k_{Ff}((s,t),(s^\prime))}{\partial l} = \alpha \times \Big[d \Big(\frac{t-t^\prime}{l}\Big) + d \Big(\frac{t^\prime-s}{l}\Big) \Big].
\end{multline}

We need the gradients of the hyperparameters of the latent function's kernel $k_{ff}$, we will not state these here as they are already well known.

For each kernel above we can compute the gradient with respect to $\alpha$ simply by returning the expression for the appropriate kernel with the initial $\alpha$ removed.

\label{multidimension}

The same idea can be used to extend the input to multiple dimensions. 
If we specify that each dimension's kernel function contains a unique lengthscale parameter, with a bracketed kernel subscript index indicating these differences, we can express the new kernel as the product of our one dimensional kernels,
\begin{equation}
k_{FF}((\bm{s},\bm{t}),(\bm{s}^\prime,\bm{t}^\prime)) = \prod_i k_{FF(i)}((s_i,t_i),(s^\prime_i,t^\prime_i)),
\label{multidim}
\end{equation}
with the cross covariance given by
$$
\mathcolorbox{yellow}{
k_{Ff}((\bm{s},\bm{t}),(\bm{t}^\prime)) = \prod_i k_{Ff(i)}((s_i,t_i),(t^\prime_i)).}
$$
\section{\hl{Non-negative latent function constraint}}
\hl{
It is common for the latent function to describe a feature which is known to be non-negative. Examples include house prices, people's heights and weights, populations, etc. We therefore may wish to constrain the model to only produce non-negative predictions. To address this problem we use as a basis the work of} \citet{riihimaki2010gaussian} \hl{who constrain a GP posterior mean to be approximately monotonic by adding `virtual points'. We could use a similar mechanism by adding virtual points that specify observations of our latent function instead. The likelihood of these new points is no longer Gaussian. Instead we use a probit function (as in the reference) with probability approaching zero if negative, and probability approaching one if positive. This non-Gaussian likelihood fails to remain conjugate with the prior. We therefore compute an approximate posterior by applying the expectation propagation (EP) algorithm, as suggested in the reference. }

\hl{We refrain from reproducing the full derivation of the EP site parameters as the full details are in }\citet{riihimaki2010gaussian}\hl{, however to summarise, we have two types of observation; integrals over the latent function and virtual observations of the latent function itself. For the former the likelihood remains Gaussian, for the latter we use a probit likelihood. The posterior is approximated using EP. We have a joint Gaussian process that describes the latent function, $f$ and its definite integrals, $F$ over hyperrectangles.} 
%Using the notation of \citet{riihimaki2010gaussian},
%
%$$
%f_{joint} = \begin{bmatrix}
%           f\\
%           F\\
%         \end{bmatrix}.
%$$
%And the covariance is described by,
%$$
%k_{joint} = \begin{bmatrix}
%           k_{ff} & k_{Ff} \\
%           k_{fF} & k_{FF} \\
%         \end{bmatrix}
%$$
\hl{We use the same expression of Bayes' rule as in }\citet{riihimaki2010gaussian},
$$
\mathcolorbox{yellow}{
p(F,f|\bm{y},\bm{z}, \bm{X}, \bm{V}) = \frac{1}{Z}p(F,f)p(\bm{y}|F,\bm{X})p(\bm{z}|f,\bm{V})}
$$
\hl{
but here $\bm{y}$ are the observations of the definite integrals (at $\bm{X}$) of the latent function, $\bm{z}$ is a placeholder vector representing the latent function's non-negative status at the virtual point locations, $\bm{V}$. The two likelihood terms are,}

$$
\mathcolorbox{yellow}{
p(\bm{y}|F,\bm{X}) = \prod_{i=1}^N \mathcal{N}\left(y_i|F(\bm{x_i}),\sigma^2\right)}
$$

$$
\mathcolorbox{yellow}{
p(\bm{z}|f,\bm{V}) = \prod_{j=1}^M \Phi\left(\frac{f(\bm{v}_j)}{\nu}\right)}
$$
\hl{The normalisation term is,}
$$
\mathcolorbox{yellow}{
Z = \int{p(F,f)p(\bm{y}|F,\bm{X})p(\bm{z}|f,\bm{V}) \; dF df}}
$$
\hl{
We then proceed with the EP algorithm to compute a Gaussian approximation to the posterior distribution,}
\begin{align*}
&\mathcolorbox{yellow}{q(F,f|\bm{y},\bm{z},\bm{X},\bm{V})} \\&\;\;\;\;\mathcolorbox{yellow}{= \frac{1}{Z_{EP}}p(F,f)p(\bm{y}|F,\bm{X})\prod_{i=1}^M t_i(\tilde{Z_i},\tilde{\mu_i},\tilde{\sigma_i})},
\end{align*}
\hl{
where $t_i$ are scaled Gaussian, local likelihood approximations, described by the three `site parameters'. Thus the posterior in this approximation is again Gaussian and a mean and covariance can be computed, using the EP algorithm (iteratively updating the site parameters and normalising term until convergence). 

A final step, once the latent function's mean and variance has been computed is to use the probit link function to generate our posterior prediction, specifically, given the distribution of the latent function prediction $p(f_*|\bm{X},\bm{y},\bm{x}_*,\bm{V})$ we produce a final prediction fed through the probit link,
$\int{\Phi(f_*) p(f_*|\bm{X},\bm{y},\bm{x}_*,\bm{V}) df_*}.$

Finally a quick note on the placement of the virtual points. The original paper discusses a few possible approaches; for low-dimensional inputs we can space these points evenly over a grid. For higher dimensions one could restrict oneself to placing these points in locations with high probability of being negative. In the examples in this paper where they are used, the dimensionality of the data set is low enough that using a grid of virtual points remains tractable.
}
\section{Arbitrary Polygon Shapes}
\label{shapes}

\hl{The product of kernels }\eqref{multidim}\hl{ assumes that we integrate between $t_i$ and $t^\prime_i$ for each dimension $i$, giving a Cartesian product of intervals. This constrains us to regions consisting of rectangles, cuboids or hyperrectangles.
Thus if our input regions are described by polytopes}\footnote{A polytope is the generalisation of a polygon to arbitrary numbers of dimensions.} that are not hyperrectangles aligned with the axes, then the above computation is less immediately tractable, as the boundaries of the integral kernels will interact. For specific cases one could envisage a change of variables, but for an arbitrary polytope we need a numerical approximation. \hl{Classical methods for quadrature (such as Simpson's method, Bayesian Quadrature, etc) are not particularly suited for this problem, either because of the potential high-dimensionality, or the non-alignment with the axes. If one considered Bayesian Quadrature} \citep{o1991bayes} \hl{for example, one is left with an analytically intractable integral, with a function with discontinuities describing the boundary of the polytope. We instead follow the more traditional approach described by }
\citet{kyriakidis2004geostatistical} \hl{who propose a numerical approximation that mirrors the exact analytical methods in this paper. Specifically they find an approximation to the double integral} \eqref{KFFa} \hl{of an underlying kernel (equation 5 in the reference). Given a uniformly random set of locations ($\bm{X}$ and $\bm{X}^\prime$) in each polygon, one sums up the covariances, $k_{ff}(\bm{x}_i,\bm{x}^\prime_i)$, for all these pairings.} Then to correct for the volumes of the two regions one divides by the number of pairings ($NN'$) and multiplies by the product of their areas/volumes ($A$ and $A^\prime$) to get an approximation to the integral,
$$
k_{FF}(\bm{X},\bm{X}') \approx \frac{A A'}{NN'} \sum_{i=1}^{N}\sum_{j=1}^{N'}{k_{ff}(\bm{x}_i, \bm{x}^\prime_j)}.
$$
Note that an advantage of this numerical approximation is the ease with which alternative kernels can be used. Their paper does not address the issue of point placement or hyperparameter optimisation. We decided the most flexible approach was to consider every object as a polytope. Each object is described by a series of $S$ simplexes, and each simplex is described by $d+1$ points (each consisting of $d$ coordinates). Selecting the simplexes is left to the user, but one could build a 3d cube (for example) by splitting each side into two triangles and connecting their three points to the cube's centre, thus forming 12 simplexes, requiring $12 \times 4 \times 3 = 144$ input values. Next, for every input polytope we place points. We summarise a method for point placement in Algorithm \ref{pickpoints} which describes how one might select points distributed uniformly within each polytope. \hl{This method guarantees points will be placed in the larger simplexes that make up the set of polytopes (if the expected number of points within that simplex is greater than one) which means that the points will be placed pseudo-uniform-randomly, aiding the approximation as this offers a form of randomised quasi-Monte Carlo sampling. We compared this to a simple Poisson-disc sampling combined with the simplex sampling to further reduce discrepancy.}\footnote{Future work might also wish to compute an equivalent to the Sobel sequence for sampling from a simplex.} Finally, for each pair of points between each pair of polytopes we compute the covariance and the gradient of the kernel with respect to the hyperparameters, $\theta$. To compute the gradient of the likelihood, $L$, with respect to the hyperparameters, we need to compute the gradients for all the $N \times N'$ point pairings, using the kernel, $k_{ff}(\cdot,\cdot)$, of the latent function, and average (taking into account the areas ($A$ and $A'$) of the two polygons);

$$
\frac{\partial L}{\partial \theta} = \frac{A A'}{N N'} \sum_{i=1}^{N} \sum_{j=1}^{N'} \frac{\partial k_{ff}(\bm{x}_i,\bm{x}^\prime_j)}{\partial \theta} \frac{\partial L}{\partial k_{ff}(\bm{x}_i,\bm{x}^\prime_j)}.
$$

\subsection{Hyperrectangle Numerical Approximation}
\label{rectangle_approximation_section}
One obvious proposal is to combine the numerical and analytical methods. We also generalise the above method to handle the covariance between a pair of sets of polytopes. Specifically, rather than approximate a set of polytopes with points, one could, conceivably achieve a higher accuracy by replacing the points with the same number of hyperrectangles, placed to efficiently fill the polytopes. As with the point method, but with hyperrectangles; we compute the covariance $k_{FF}$ between all pairings of hyperrectangles from the different sets of polytopes and then sum these to produce an estimate for the covariance between the two sets of polytopes (potentially correcting for the volume of the two sets of polytopes if the two sets of hyperrectangles do not completely fill them). Specifically, we compute,
$$
k_{FF}(\bm{X},\bm{X}') \approx \sum_{i=1}^{N}\sum_{j=1}^{N'}\frac{A_i A'_j}{a_i a'_j}{k_{FF}(\bm{x}_i, \bm{x}^\prime_j)},
$$
where $A_i$ refers to the volume of the polytope associated with hyperrectangle $i$ (note other hyperrectangles may also be associated with that polytope), and $a_i$ is the sum of the volumes of all the hyperrectangles being used to approximate the same polytope. Thus their ratio gives us a correction for the hyperrectangle's volume shortfall.

\newcommand*\Let[2]{\State #1 $\gets$ #2}
\begin{algorithm}
  \newcommand\mdoubleplus{\mathbin{+\mkern-10mu+}}
  \newcommand{\Polytope}{\mathcal{T}}
  \newcommand{\Simplex}{\mathcal{S}}
  \newcommand{\Points}{P}  
  \begin{algorithmic}[1]
    \Require{$\Polytope$, the polytope we want to fill with samples - described by a list of $d \times n$ matrices defining simplexes. $d$ spatial dimensions and $n=d+1$ vertices.}
    \Require{$\rho$, density of points (points per unit volume)}
    \Statex
    \Function{GetUniformSamples}{$\Polytope$, $\rho$}
      \For {\textsc{Simplex}, $\Simplex$ \texttt{in} $\Polytope$}
        \Let{$V$}{$\textsc{CalcVolume}(\Simplex)$}
        \For {$0 \leq i < V\rho$}
          \Let{$\Points$}{$\Points \cup \textsc{SimplexRandomPoint}(\Simplex)$}
        \EndFor
      \EndFor
    \EndFunction
    \\
    
    \Function{CalcVolume}{$\Simplex$ made of vertices $\bm{v}_0 ... \bm{v}_{n-1}$}
    
    \Comment{modified from \citet{stein1966note}}
    
		%Rows of matrix $M$ are vertex coordinates relative to 0th vertex.
    	%$(m_{ij}) \in \mathbb{R}^{d \times d}$ 
    	%\Let{$M_{i:}$}{$v_{i+1} - v_0$} 
    	
    	%\Return $\frac{1}{d!}{|\det[M]|}$ 
        \Return $\left|{1\over {d!}}\det \left[ \bm{v}_1-\bm{v}_0, \bm{v}_2-\bm{v}_0, \dots, \bm{v}_{n-1}-\bm{v}_0 \right] \right|$

    \EndFunction
    \\
    
    \Function{SimplexRandomPoint}{$\Simplex$} 

    \Comment{Algorithm duplicated from \citet{grimme2015picking}}
    	\Let{z}{$[1] \mdoubleplus \texttt{uniform}(d) \mdoubleplus[0]$}\Comment{see footnote{${}^\dagger$}}
    	\Let{$l_i$}{$z_i^{1/(n-i)}\;\;\;\;1\leq i \leq n$}

    	\Return {$\sum_{i=1}^{n}{(1-l_i)(\prod_{j=1}^{i}{l_j})\bm{v}_i}$}
    \EndFunction
  \end{algorithmic}
  \caption{Pick a random point inside a polytope.}
  {${}^{\dagger}$\footnotesize uniform($d$) selects $d$ uniformly random numbers. $\mdoubleplus$ is the concatenation operator.}
  \label{pickpoints}
\end{algorithm}
The placement of the hyperrectangles is a more complex issue than the placement of the points in the previous section. For the purposes of this paper we use a simple greedy algorithm for demonstration purposes. Other work exists on the time complexity and efficient placement of rectangles to fill a polygon, although many either allow the rectangles to be non-axis-aligned or requires the polygon to be an L shape \citep{iacob2003covering} or orthogonal, or are only for a single rectangle \citep{daniels1997finding} in a convex polygon \citep[e.g.][]{knauer2012largest,alt1995computing,cabello2016finding}. We found the straightforward greedy algorithm to be sufficient.

%removed this ref.
%%@article{levcopouloslinear,
%  title={A Linear-Time Approximation Algorithm for Minimum Rectangular Covering},
%  author={Levcopoulos, Christos and Gudmundsson, Joachim}
%}

\section{Results}
\hl{We illustrate and assess the above methods through a series of experiments. We start, in Section }\ref{robotexample} \hl{with a simple one-dimensional example in which we have noisy observations of a series of definite integrals and we want to estimate the latent function. In Section} \ref{nonnegconst} \hl{we use another synthetic dataset to illustrate the non-negative virtual point constraints on the posterior. In Section} \ref{diffpriv} \hl{we use a real dataset describing the age distribution of a census tract, with the individuals providing the data made private through the differential privacy framework} \citep{dwork2014algorithmic}. \hl{We demonstrate how the method can support inference on noisy, differentially private data and test the non-negative constrained integral. In Section }\ref{section_citibike} \hl{we consider another histogram example, but this time with a higher dimensional input, of the durations of hire bike users, given the start and finish station locations. In Section} \ref{audience} \hl{we extend the method to predict other integrals (not just densities). Finally in Section} \ref{section_popden} \hl{we consider non-rectangular input volumes and compare numerical approximations for GP regression. In these later sections the latent function output is far from zero, thus the non-negative constraint had no effect (and is not reported).}

\subsection{Speed Integration Example}
\label{robotexample}
Before looking at a real data example, we illustrate the kernel with a simple toy example. We want to infer the speed of a robot that is travelling along a straight line. The distance it has travelled between various time points has been observed, as in Table \ref{robot_table}. A question we might ask, how fast was the robot moving at $5$ seconds? We enter as inputs the four integrals. We select the lengthscale, kernel variance and Gaussian noise scale by maximising the log marginal likelihood, using gradient descent \citep[Section 5.4.1]{williams2006gaussian}. We now can make a prediction of the latent function at five seconds using standard GP regression. Specifically the posterior mean and variances are computed to be,
\begin{align}
\bar{f_*} &= \bm{k}_{F*}^\top (\bm{K}_{FF} + \sigma^2 I)^{-1} \bm{y}\\
\textsc{V}[f_*] &= k_{**} - \bm{k}_{F*}^\top (\bm{K}_{FF} + \sigma^2 I)^{-1} \bm{k}_{F*},
\end{align}
where $\bm{K}_{FF}$ is the covariance between pairs of integrals, $\bm{k}_{F*}$ is the covariance between a test point in latent space and an integral. $\sigma^2$ is the model's Gaussian noise variance. $\bm{y}$ are the observed integral outputs and $k_{**}$ is the variance for the latent function at the test point.

\hl{The optimal hyperparameters that maximise the log marginal likelihood, are for the kernel to have variance of 12.9$\text{m}^2 \text{s}^{-2}$ and lengthscale 7.1s, model likelihood Gaussian noise, 0.6$\text{m}^2 \text{s}^{-2}$.} %The relatively long lengthscale means the speed increases roughly linearly. One can see that this is appropriate for the observations made, in which the rate that distance is covered increases linearly with time.

\begin{figure}[tb]
  \begin{center}
  \includegraphics[width=0.5\textwidth]{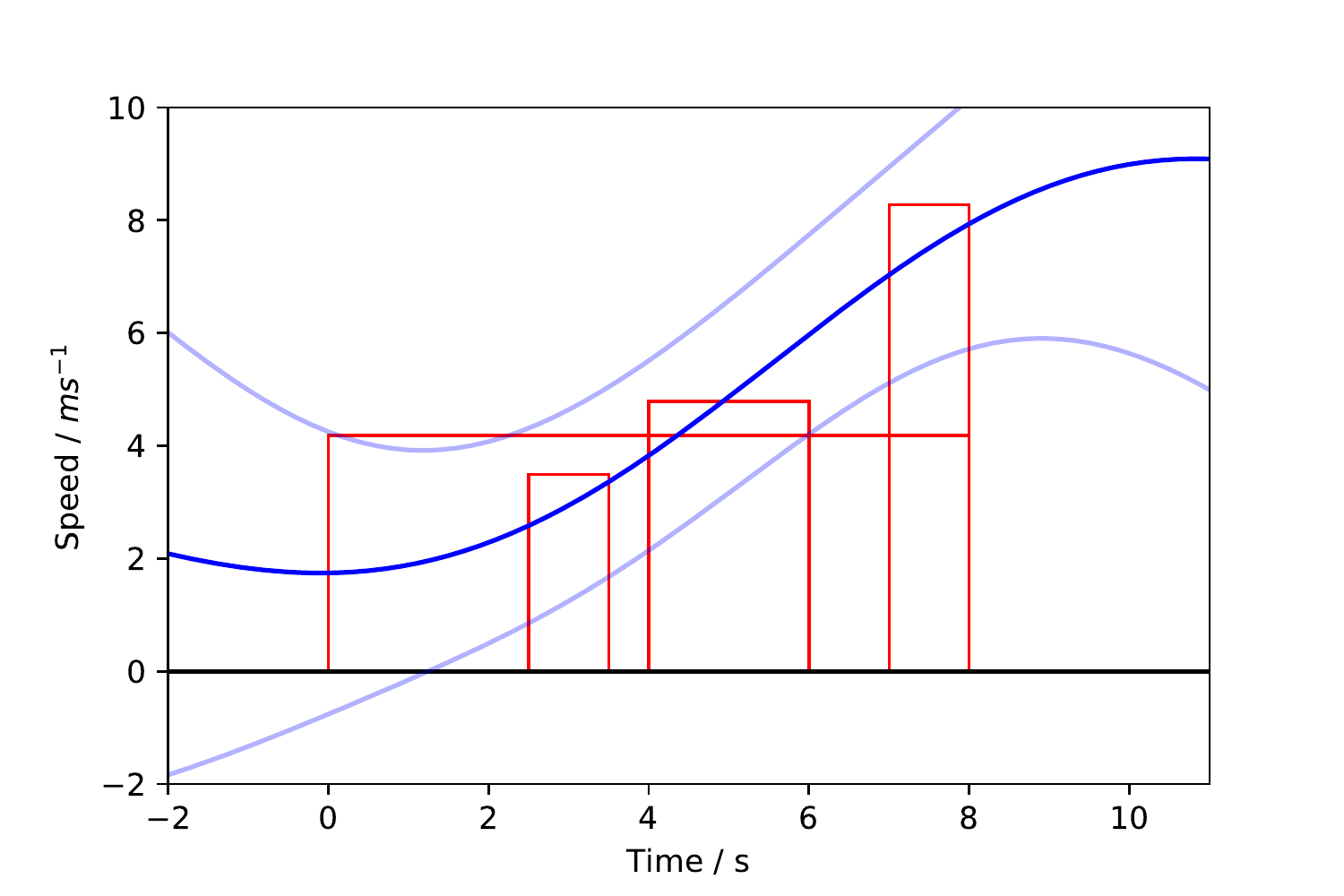}
  \end{center}
\caption{\hl{Illustration of how the robot's speed can be inferred from a series of observations of its change in location, here represented by the areas of the four rectangles. The blue lines indicate the posterior mean prediction and its 95\% confidence intervals.}}  
  \label{robot}
\end{figure}

Figure \ref{robot} illustrates the four observations as the areas under the four rectangles, and shows the posterior prediction of the GP.
\hl{To answer the specific question above, the speed at $t=5\text{s}$ is estimated to be $4.87 \pm 1.70 \text{m} \text{s}^{-1}$ (95\% CI). We constructed the synthetic data with a function that increases linearly at $1 \text{ms}^{-2}$, with added noise. So the correct value lies inside the prediction's CIs.}

\begin{table}
\begin{center}

\begin{tabular}{l l l}
Start location / m & End location / m & Time \\
\hline       
0 & 8 & 33.47 \\
2.5 & 3.5 & 3.49 \\
4 & 6 & 9.56 \\
7 & 8 & 8.27 \\
\hline
\end{tabular}
\end{center}
\caption{\hl{Simulated observations of robot travel distances. Figure} \ref{robot} \hl{illustrates these observations with rectangle areas.}}
\label{robot_table}
\end{table}
\subsection{\hl{Non-negative constraint}}
\label{nonnegconst}
\begin{figure}[tb]
  \begin{center}
  \includegraphics[width=0.5\textwidth]{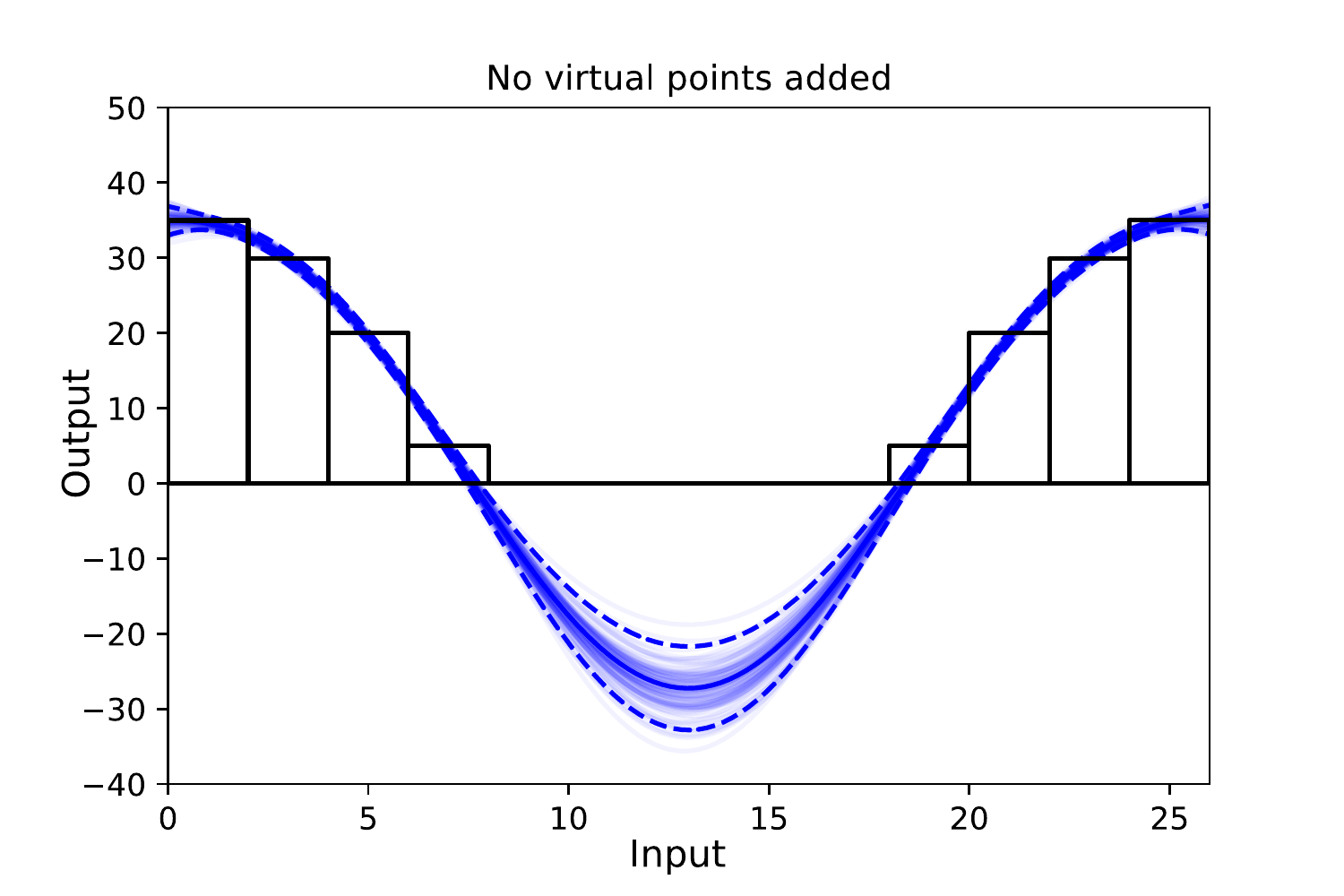}
  \includegraphics[width=0.5\textwidth]{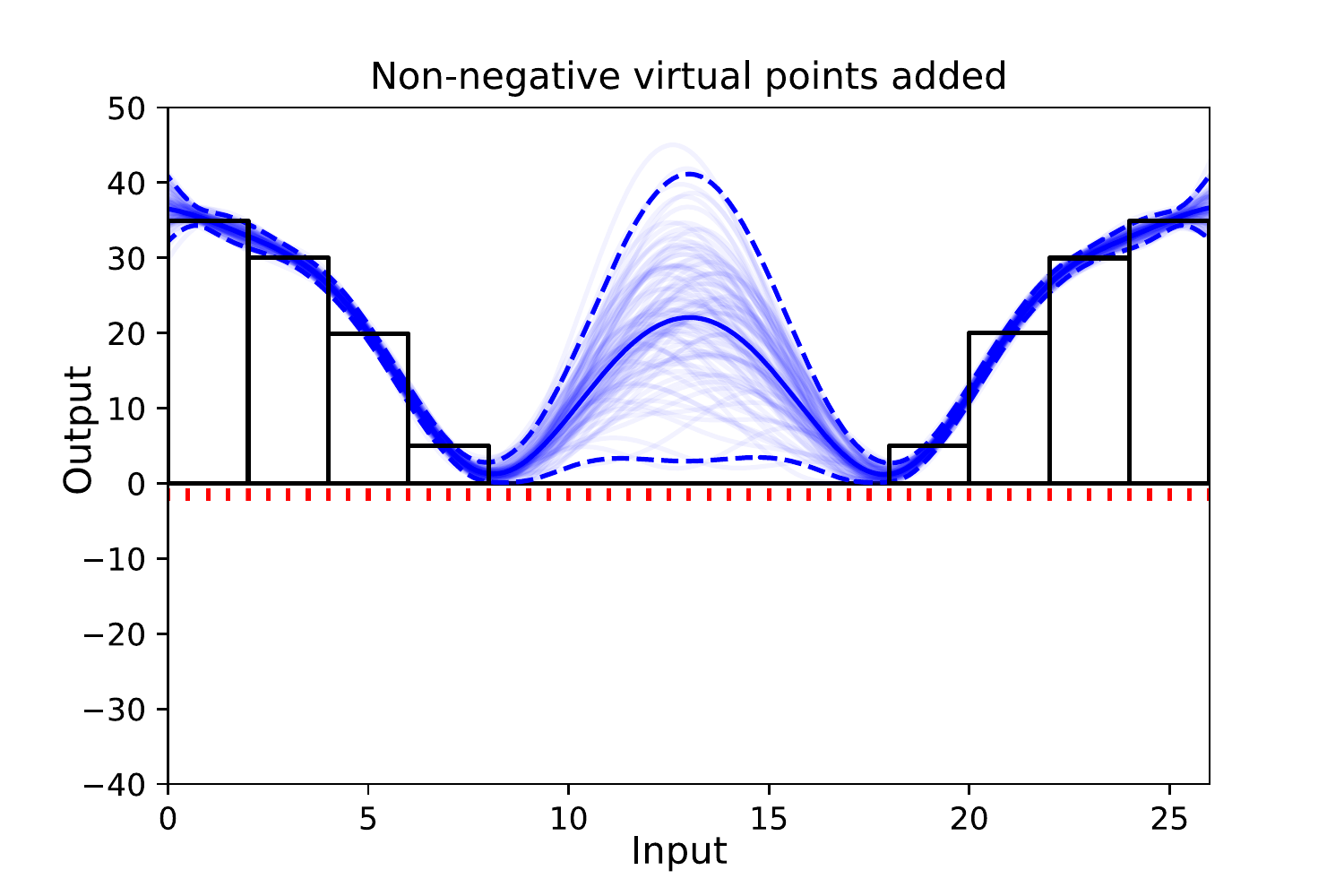}  
  \end{center}
\caption{\hl{Synthetic dataset demonstrating the use of virtual points (locations indicated by red ticks below axis) to enforce non-negativity. Mean, solid blue line; 95\%-CI, dashed blue line. The upper figure uses a simple Gaussian likelihood without virtual points. The lower figure has a grid of virtual points and a probit likelihood function for these observations. Note that the latent posterior mean and uncertainty is fed through this link function to produce the mean and CIs plotted.}}
  \label{virtual}
\end{figure}

\hl{As a simple demonstration of the non-negative constraint in operation, we consider a synthetic one dimensional dataset of eight observations arranged to encourage the posterior mean to have a negative region. We then place a regular grid of fifty-three virtual points over the domain. Figure }\ref{virtual}\hl{ illustrates both the standard Gaussian-likelihood prediction and the result with these probit-likelihood virtual points. There are no observations between output locations eight and eighteen leaving the function largely unconstrained thus there is large uncertainty in this part of the domain. The reader may notice that the uncertainty in this part of the domain is greater for our constrained model. This is not directly a result of the constraints, but rather due to shorter lengthscales. When the GP hyperparameters were optimised for the constrained example, the lengthscales chosen by the ML procedure were significantly shorter (4.36 instead of 10.37), one can see that this is necessary, as any function that both fits the data but also avoids becoming negative requires a relatively steep change in gradient (around eight and eighteen in the plot).}

\subsection{Differentially Private Age Data}
\label{diffpriv}
We consider the age distribution of 255 people from a single output area (E00172420) from the 2011 UK census.\footnote{a peak of students at age 18 was removed, so the graph only includes the permanent residents of the area.} We also make this histogram differentially private, to demonstrate the improved noise immunity of the new method. We group the people into a histogram with equal ten year wide bins, and add differentially private noise using the Laplace mechanism \citep[][section 3.3]{dwork2014algorithmic}. Specifically we take samples from a scaled Laplace distribution and add these samples to the histogram's values. The Laplace noise is scaled such that the presence or absence of an individual is provably difficult to detect, using the $\varepsilon$-DP Laplace mechanism. One can increase the scale of the noise (by reducing $\varepsilon$) to make it more private, or increase $\varepsilon$, sacrificing privacy for greater accuracy. The aim is to predict the number of people of a particular age. We use four methods; \hl{(i) simply reading off the bin-heights, (ii) fitting a standard GP (with an EQ kernel) to the bin centroids, (iii) using a GP with the integral kernel or (iv) using the integral kernel, constrained to be non-negative.}

\begin{figure}[tb]
  \begin{center}
  \includegraphics[width=0.5\textwidth]{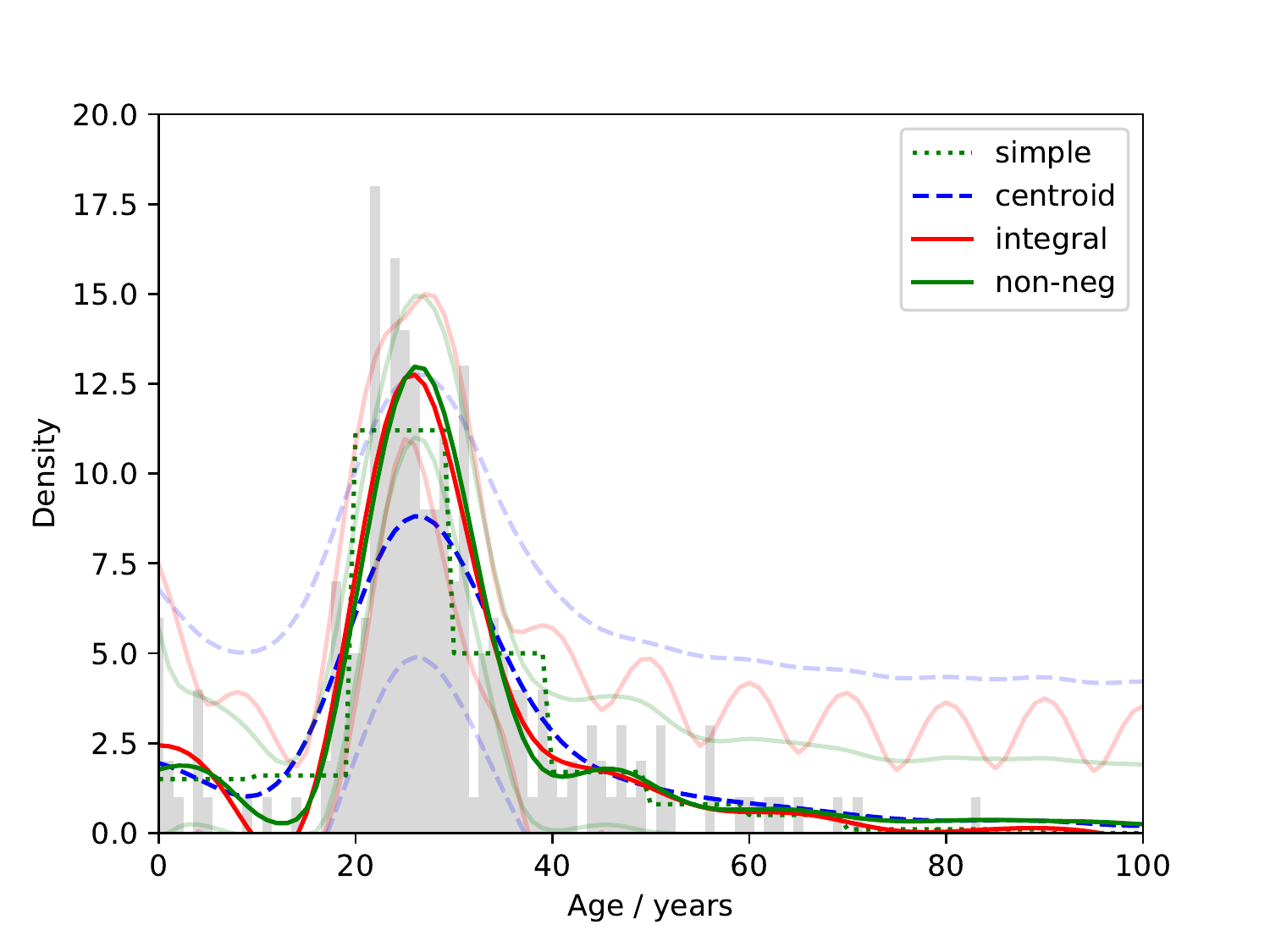}
  \end{center}
\caption{\hl{Various fits to an age histogram of 255 people in which the data has been aggregated into ten-year wide bins (in this case we have not added DP noise), with the original data plotted as grey bars. The dotted, green line uses the bin heights to make predictions directly. The dashed, blue line fits an EQ GP to the centroids of the bins. The red solid line is the prediction using the integral kernel. The solid green line is the integral kernel constrained to be positive. The 95\% confidence intervals are indicated with fainter lines.}}  
  \label{int_kernel}
\end{figure}
\begin{table*}
\begin{center}
\begin{tabular}{l r r r r r }
\hline
\hline$\varepsilon$  & Simple  & Centroid  & Integral  & Non-Neg & 95\% CI\\
\hline0.01 & $15.505$ & $6.165$ [$16\%$] & $5.236$ [$7\%$] & $5.282$ [$54\%$]& 0.024\\
0.10 & $5.242$ & $2.139$ [$25\%$] & $2.158$ [$9\%$] & $2.826$ [$35\%$]& 0.051\\
0.20 & $5.087$ & $1.916$ [$22\%$] & $1.735$ [$8\%$] & $1.966$ [$12\%$]& 0.046\\
0.50 & $5.032$ & $1.874$ [$19\%$] & $1.652$ [$7\%$] & $1.715$ [$11\%$]& 0.012\\
1.00 & $5.033$ & $1.835$ [$16\%$] & $1.611$ [$7\%$] & $1.696$ [$12\%$]& 0.005\\
not-DP & $5.030$ & $1.962$ [$7\%$] & $1.604$ [$7\%$] & $1.690$ [$12\%$]& 0.000\\\hline
\end{tabular}

%\begin{tabular}{l l l l r}
%$\varepsilon$ & Simple & Centroid & Integral & 95\% CI\\
%\hline
%$ 0.05 $ & $ 5.86 $ & $ 3.11 $ & $ 3.04 $ & 0.06\\
%$ 0.1 $ & $ 5.36 $ & $ 2.45 $ & $ 2.19 $ & 0.03\\
%$ 0.2 $ & $ 5.21 $ & $ 2.07 $ & $ 1.78 $ & 0.03\\
%$ 0.5 $ & $ 5.17 $ & $ 1.90 $ & $ 1.65 $ & 0.02\\
%$ 1.0 $ & $ 5.16 $ & $ 1.83 $ & $ 1.63 $ & 0.01\\
%$ 2.0 $ & $ 5.16 $ & $ 1.78 $ & $ 1.63 $ & 0.01\\
%$ 5.0 $ & $ 5.16 $ & $ 1.77 $ & $ 1.63 $ & $ < 0.01$\\
%$ \infty $ & $ 5.16 $ & $ 1.76 $ & $ 1.63 $\\
%\hline
%\end{tabular}

%\begin{tabular}{l r r r }
%$\varepsilon$ & Simple  & Centroid  & Integral \\
%\hline
%0.05 & 5.84 & 3.10 & 3.00\\
%0.1 & 5.34 & 2.42 & 2.13\\
%0.2 & 5.21 & 2.13 & 1.77\\
%0.5 & 5.17 & 1.90 & 1.65\\
%1.0 & 5.16 & 1.79 & 1.63\\
%2.0 & 5.16 & 1.79 & 1.63\\
%5.0 & 5.16 & 1.76 & 1.63\\
%$\infty$ & 5.16 & 1.76 & 1.63\\
%\hline
%\end{tabular}
\end{center}
\caption{\hl{RMSE for all 100 age bins, for the simple (directly read off histogram), centroid (EQ GP fit to bin centres), integral method and the integral method with the non-negative constraint, for various levels of differential privacy. Computed RMSE using 30 DP samples. 10,000 bootstrap resamples used to compute 95\% CI estimate (to 1 significant figure), value quoted is largest of four columns for simplicity. In [brackets] we have recorded the percentage of predictions that lay outside the 95\% CI of the posterior. Bin size, 10 years.}}
\label{age_table}
\end{table*}
Figure \ref{int_kernel} demonstrates these results. Note that the GP with an integral kernel will attempt to model the area of the histogram, leading to a more accurate prediction around the peak in the dataset. \hl{The figure also indicates the uncertainty quantification capabilities provided by using a GP. Not all applications require or will use this uncertainty, but we have briefly quantified the accuracy of the uncertainty by reporting in Table} \ref{age_table} \hl{the proportion of the original training data that lies outside the 95\% CI (one would expect, ideally, that this should be about 5\%).}

\hl{To explore the interaction of the methods with the addition of noise to the data, we manipulate the scale of the DP noise (effectively increasing or decreasing the scale of the Laplace distribution) and investigate the effect on the RMSE of the four methods. Remember that decreasing the value of $\varepsilon$ makes the prediction more private (but more noisy). This is not cross-validated leave-one-out, as the use-case includes the test point in the aggregation.}

Table \ref{age_table} illustrates the effect of the DP noise scale on the RMSE. We find that the integral method performs better than the others for all noise-scales tested. Intriguingly the simple method seems to be less affected by the addition of DP noise, possibly as the two GP methods effectively try to use some aspect of the gradient of the function, an operation which is vulnerable to the addition of noise. \hl{The integral method seems particular useful in the most commonly used values of $\varepsilon$, with meaningful reductions in the RMSE of 13\%. The non-negative method does not perform as well. We have set to zero all the negative training data (by default adding the DP noise will make some of the training points negative). If one considers the portion of the domain over 60 years, one can see that the mean of the non-negative-constrained integral-kernel posterior is a little above the others. This occurs as, if one imagines the effect of the non-negative constraint on all possible functions, they will all lie above zero, thus the constraint pushes the mean upwards, i.e. even if the mean without the constraint was non-negative, this mean would have included negative examples. The effect is a worsening of the RMSE/MAE, as many of the training points are zero (which is now further from the posterior mean). The proportion that fall inside the 95\% CI is also low as many of the test points are zero, and this model's 95\% CI typically will not quite include zero itself.}

\hl{To describe the DP noise (and potentially differences in the expected noise if the integrals represent means, for example) we added a white noise heteroscedastic kernel to our integral kernel. This effectively allows the expected Gaussian noise for each observation to be specified separately. One could for example then specify the Gaussian noise variance as $\sigma^2/n_i$ where $n_i$ is the number of training points in that histogram bin (if finding the mean of each bin). If the histogram is the sum of the number of items in each bin we should use a constant noise variance.}

\subsection{Citibike Data (4d hyperrectangles)}
\label{section_citibike}
The above example was for a one-dimensional dataset. We now consider a 4d histogram. The New York based citibike hire scheme provides data on the activities of its users. Here we use the start and end locations (lat and long) to provide the four input dimensions, and try to predict journey duration as the output. To demonstrate the integral kernel we bin these data into a 4-dimensional grid, and find the mean of the data points that lie within each bin. To investigate the integral kernel's benefits we vary the number of bins and the number of samples in these bins. As before we compare against an alternative method in which we fit a GP (using an EQ kernel) to the centroids of the bins. \hl{Note that bins that contain no datapoints were not included as training data (as their mean was undetermined). We chose the two models' hyperparameters using a grid search on another sample of citibike data and assess the models by their ability to predict the individual training points that went into the aggregation.}

Table \ref{citibike_table} illustrates these results. \hl{One can see obvious features; more samples leads to more accurate predictions and small numbers of bins causes degradation in the prediction accuracy. However, most interesting is how these interact with the two methods. We can see that for low numbers of bins the integral method does better than the centroid method. With, for example, $5^4 = 625$ bins, the integral method provides no additional support as the data is spread so thinly amongst the bins. The integral kernel will simply act much like the latent EQ kernel. Specifically, these first two experiments suggest that when there are many data points the two methods are fairly comparable, but the integral kernel is of greatest utility when there either are few samples (as shown in Table} \ref{citibike_table}\hl{), or they contain considerable noise (as shown in Table} \ref{age_table}\hl{, for low values of $\varepsilon$).}

\begin{table*}
\begin{center}
\begin{tabular}{l r r r r r r r r r r }\\
\hline
 & \multicolumn{10}{c}{Number of bins}\\
 & \multicolumn{2}{c}{$2^4$}  & \multicolumn{2}{c}{$3^4$}  & \multicolumn{2}{c}{$4^4$}  & \multicolumn{2}{c}{$5^4$}  & \multicolumn{2}{c}{$6^4$} \\
Samples  & Centroid & Integral  & Centroid & Integral  & Centroid & Integral  & Centroid & Integral  & Centroid & Integral \\
\hline
80 & {487.4} & \textbf{465.1} & {494.9} & \textbf{482.8} & {485.9} & \textbf{469.9} & {478.5} & \textbf{452.8} & \textbf{471.3} & {474.7} \\
160 & {487.4} & \textbf{475.5} & \textbf{471.1} & {481.6} & \textbf{451.0} & {464.2} & \textbf{425.5} & {435.7} & \textbf{406.9} & {408.9} \\
320 & {473.1} & \textbf{466.2} & {453.4} & \textbf{438.3} & {422.4} & \textbf{388.9} & \textbf{408.9} & {431.9} & {378.8} & \textbf{374.0} \\
640 & {468.6} & \textbf{465.5} & {449.5} & \textbf{435.5} & {413.4} & \textbf{374.7} & \textbf{372.5} & {379.9} & {366.7} & \textbf{365.2} \\
1280 & {469.8} & \textbf{467.5} & {450.7} & \textbf{446.3} & {418.1} & \textbf{375.2} & \textbf{382.1} & {385.3} & {375.6} & \textbf{370.1} \\
2560 & {468.9} & \textbf{467.7} & {456.9} & \textbf{436.4} & {420.3} & \textbf{373.1} & {363.1} & \textbf{360.3} & \textbf{353.0} & {359.3} \\
\hline
\end{tabular}

%\begin{tabular}{l r r r r r r r r }
%\hline
% & \multicolumn{8}{c}{Number of bins}\\
%  & \multicolumn{2}{c}{${1^4}$} & \multicolumn{2}{c}{${3^4}$} & \multicolumn{2}{c}{${7^4}$} & \multicolumn{2}{c}{${15^4}$}\\
%Samples & centroid & integral & centroid & integral & centroid & integral & centroid & integral\\
%\hline
%160 & 495.6 & 491.3 & 438.1 & 435.5 & 524.6 & 452.6 & 739.4 & 489.1\\
%320 & 493.5 & 490.9 & 441.7 & 433.6 & 414.7 & 381.8 & 590.2 & 481.5\\
%640 & 497.0 & 496.7 & 443.9 & 410.4 & 401.6 & 383.8 & 565.0 & 479.4\\
%1280 & 497.6 & 497.2 & 438.0 & 396.0 & 381.1 & 373.5 & 513.5 & 473.5\\
%2560 & 497.8 & 497.4 & 444.8 & 399.3 & 369.4 & 361.0 & 460.7 & 456.7\\
%\hline
%\end{tabular}
\end{center}
\caption{\hl{Mean Absolute Error in predictions of journey duration (in seconds) for the citibike dataset using the integral and centroid methods, over a variety of sample counts and bin counts. 1000 randomly chosen journeys were used in the test set, experiment performed once for each configuration. Bold highlights best of each pair.}}
\label{citibike_table}
\end{table*}

\subsection{Audience Size Estimation (predicting integrals not densities)}
\label{audience}
\begin{figure*}[tb]
  \begin{center}
  \includegraphics[width=0.8\textwidth]{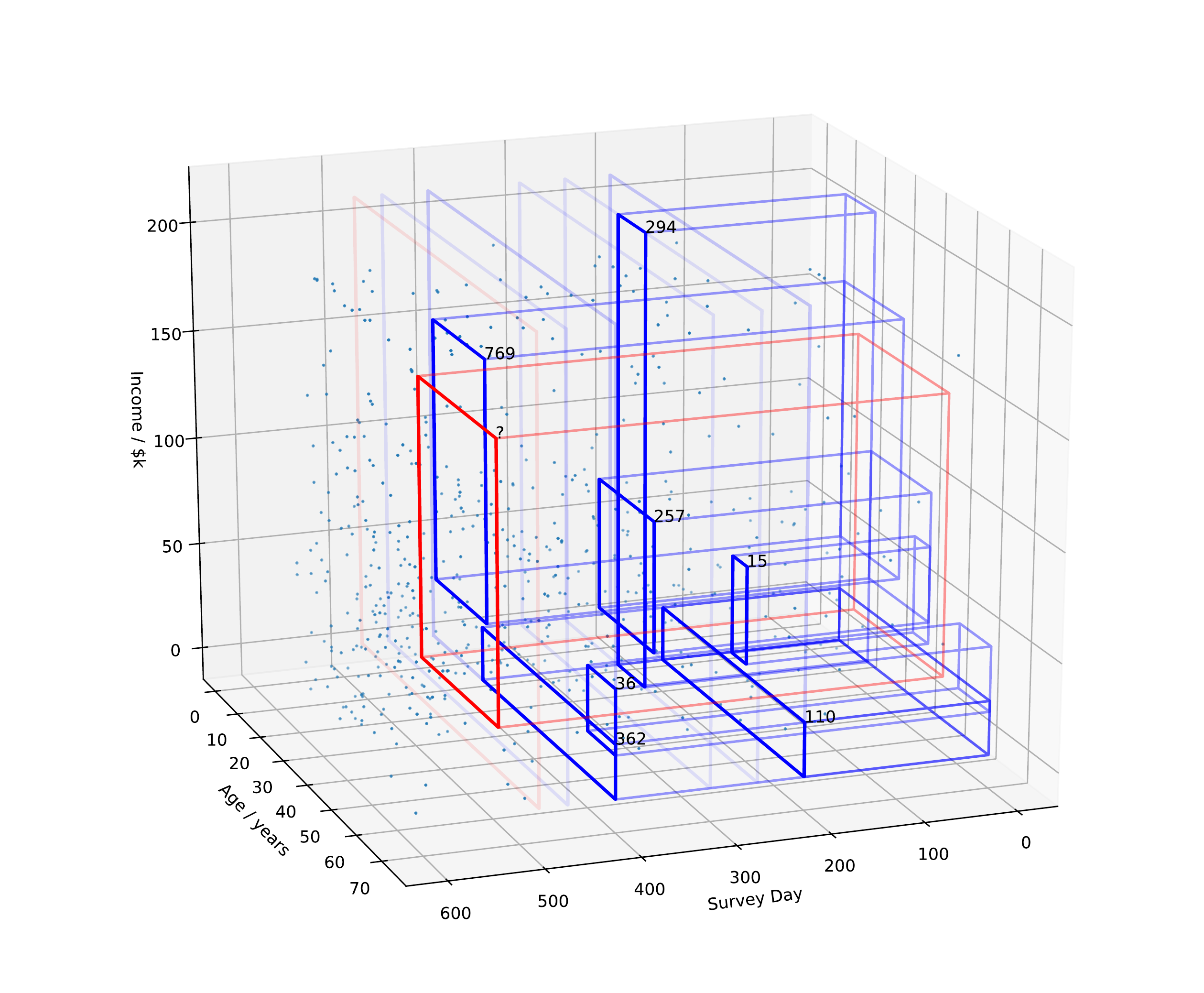}
  \end{center}
  \vspace{-10mm}
\caption{\hl{A demonstration of the audience-size estimation problem. Within the 3d volume lie the individuals that make up the population subscribed by the company. Their location in 3d specified by the date they joined, their income and age. Seven previous surveys (in blue) have been performed over a growing group of clients. Each survey is indicated by a rectangle to indicate the date it occurred and the age/income of participants recruited. All the participants within the cuboid projected backwards from the rectangle are those that had already registered by the date of the survey and so could have taken part. Each volume is labelled with the number of people which took part in each survey. In red is a new survey we want to estimate the count for.}}
  \label{audiencedemo}
\end{figure*}

\begin{table}
\begin{center}
\begin{tabular}{l r l r l}
\hline
 & \multicolumn{4}{c}{Method} \\
 & \multicolumn{2}{c}{Integral} & \multicolumn{2}{c}{Centroid} \\
\hline
RMSE & 126.3 & $\pm$ 10.74 & 223.1 & $\pm$ 14.88 \\
MAE & 73.1 & $\pm$ 4.52 & 143.5 & $\pm$ 7.51 \\
\hline
\end{tabular}
\end{center}
\caption{RMSE and MAE for 1998 randomly generated audience survey requests. 95\% CIs for these statistics was calculated using non-parametric Monte Carlo bootstrapping with 100,000 samples with replacement.}
\label{audience_table}
\end{table}

We may also wish to produce predictions for new bins. A motivating, real, example is as follows. Imagine you work for a market research company and have a cohort of citizens responding to your surveys. Each survey requires a particular audience answers it. For example the company's first survey, in January, required that respondents were aged between 30 and 40 of any income. They had 320 replies. In February a second survey required that the respondents were aged between 25 and 35 and earned at least \$30k, and had 210 replies. In March their third survey targeted those aged 20 to 30 with an income less than \$40k. How many respondents might they expect? The latent function is population density across the age and income axes, while the outputs are population counts. \hl{We can use expressions} \eqref{kFF}~\&~\eqref{kFf} as described at the start of Section \ref{analytical_derivation} but use $k_{FF}$ instead of $k_{Ff}$ when making predictions (the inputs at the test points now consist of the boundaries of an integral, and not just the location to predict a single density). Figure \ref{audiencedemo} illustrates this with a fictitious set of surveys targeting sub-groups of the population.

We simulated a population of 5802 survey-takers by sampling from the US census bureau's 2016 family income database. We distribute the start dates randomly, with a skew towards younger participants. For the example in Figure \ref{audiencedemo} we computed a prediction for the test region using the integral kernel. We compared this to a model in which the counts had been divided by the volumes of the cuboids to estimate the density in each, and used these with the centroid locations to fit a normal GP (with an EQ kernel) to estimate the density (and hence count) in the test cuboid. For this case we found that both methods underestimated the actual count (of 1641). The centroid method predicted 1263 (95\% CI: 991-1535), while \hl{the integral} method predicted 1363 (95\% CI: 1178-1548). The shortfalls are probably due to the skew in the participant start times towards the older portion. The previous training cuboids would have had lower densities, leading to the underestimates here. Intriguingly the integral method still produces a more accurate prediction.

To test this more thoroughly, we simulate 1998 sets of surveys (between 6 and 19 surveys in each set) over this data, and compare the RMSE (and MAE) of the two methods when predicting the number of respondents to a new survey. Table \ref{audience_table} shows that the integral method produces more accurate results in this simulated dataset. 

\subsection{Population Density estimates (2d non-rectangular disjoint inputs)}
\label{section_popden}
\begin{figure}[tb]
  \begin{center}
  \includegraphics[trim=1.5cm 1.5cm 1.5cm 1.5cm,clip,width=0.5\textwidth]{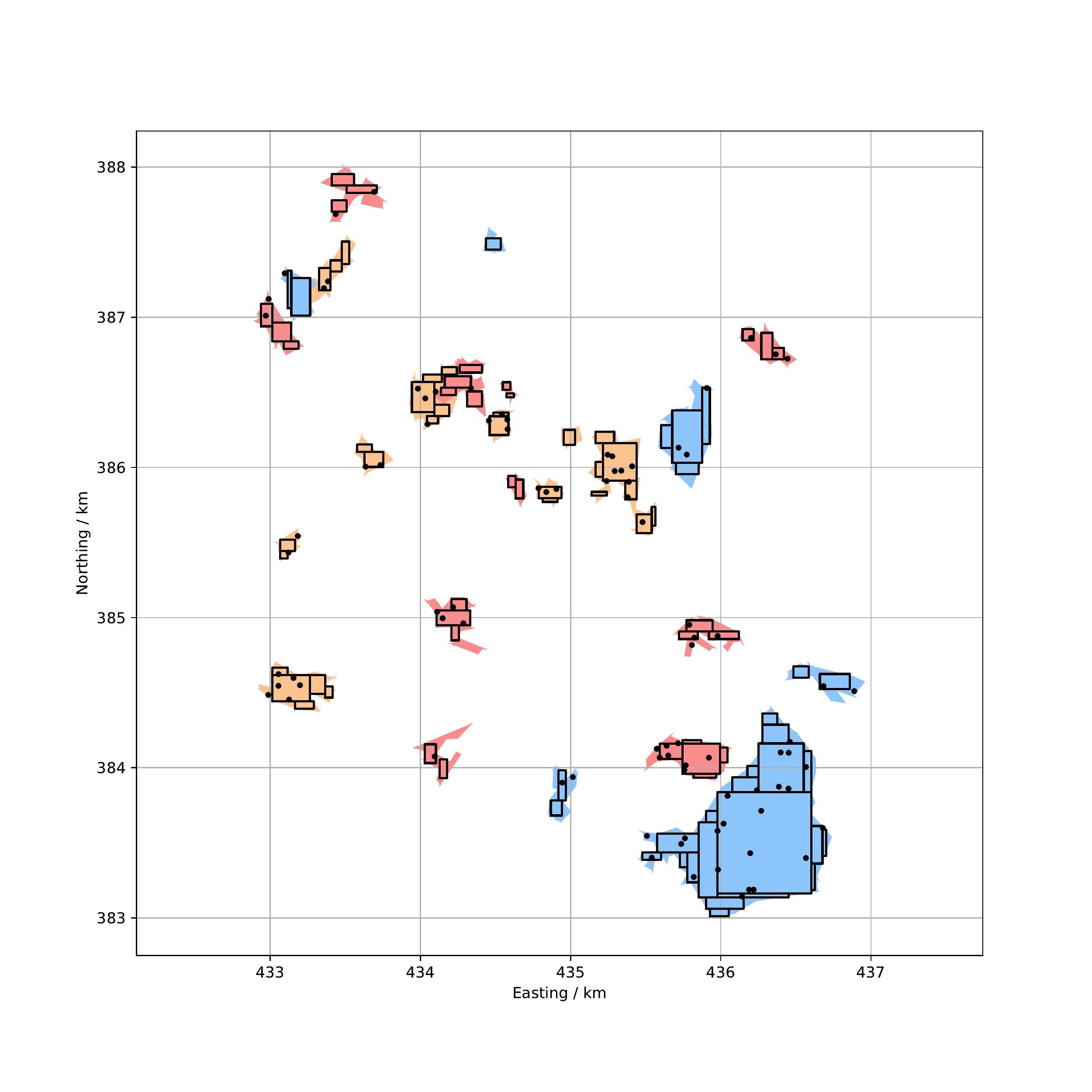}
  \end{center}
\caption{Example of both rectangular and point approximation to three sets of polygons (from the census output areas of Sheffield). With approximately 30 rectangles or points used for each set.}
  \label{rectangle_example}
\end{figure}

\begin{table}
\begin{center}

\begin{tabular}{l c c c c }
\hline
Number of & & & & \\
approximating & \multicolumn{3}{c}{Mean Abs. Error} & Std. \\
points & points & low-disc & hyperrects. & error \\
\hline
2 & 242.3 & 242.0 & 198.0 & 2.1 \\
4 & 209.2 & 207.7 & 183.8 & 2.5 \\
8 & 194.4 & 193.6 & 185.1 & 1.8 \\
16 & 187.3 & 189.4 & 187.1 & 1.1 \\
32 & 186.1 & 187.6 & 185.7 & 0.9 \\
64 & 185.5 & 185.9 & 185.4 & 0.4 \\
\hline
\end{tabular}

\end{center}
\caption{\hl{Number of integration approximation features per input for points, lower-discrepancy points and hyperrectangle shape integral methods, and the effect this has on the MAE of the output area density predictions (population density, people per $\text{km}^2$). Reported MAE based on average of twenty point placement iterations. Maximum std. error for each row shown (computed from 14 runs of each). Lengthscale = 160m. kernel variance = 160. Gaussian likelihood variance = 1, variances originally in units of people${}^2$ but the outputs were normalised.} % variance = 20 people${}^2$ (normalised), Gaussian noise variance = $10^{-6}$ people${}^2$  (normalised).
}
\label{pointeffect}
\end{table}

In earlier sections we assumed rectangular or cuboid input regions in the training set. However many datasets contain more complicated shapes. In this section we briefly apply the numerical approximation devised by \citet{kyriakidis2004geostatistical} and extended in section \ref{shapes} \hl{to use hyperrectangles to fill the polytopes}. In this example we use the population density of areas from the UK census. \hl{In particular those output areas lying within a} $16 \text{km}^2$ square, centred at Easting/Northing 435/386 km (Sheffield, UK). We assume, for this demonstration, that we are given the total population of a series of 40 groupings of these output areas (the output areas have been allocated to these sets uniformly and randomly). This simulates a common situation in which we know the aggregate of various subpopulations. The task then is to predict the population density of the individual output areas that make up the aggregates. Figure \ref{rectangle_example} demonstrates example placement results, while Table \ref{pointeffect} \hl{demonstrates the effect of changing the number of points/rectangles on the MAE. For the lowest numbers of approximating points the hyperrectangle approximation had a lower MAE for an equal number of approximating points. Significantly more points were needed (approximately 3-4 times as many) when using points to approximate the MC integration than when using the hyperrectangles, to reach the same accuracy. 
The lower-discrepancy sampling did not appear to significantly improve the results of the point approximation.}

\hl{As another example we look at the covariance computed between three sets of polygons illustrated in Figure} \ref{rectangle_example}. \hl{We test both the point- and hyperrectangle- approximations. Table} \ref{rectangle_table} \hl{shows the MAE when computing the covariance between these three sets of polygons. Using the rectangle approximation reduces the error by approximately 4 times, for the same number of training points/rectangles.}

\begin{table}[H]
\begin{center}
\begin{tabular}{l l l }
\hline
 Number of points & \multicolumn{2}{c}{Mean Abs Error} \\
 or rectangles & Points & Rectangles\\
\hline
16 & 0.0197 & 0.0049 \\
32 & 0.0084 & 0.0018 \\
64 & 0.0007 & 0.0002 \\
128 & 0.0004 & $<0.00015$ \\
\hline
\end{tabular}
\end{center}

\caption{Mean Absolute Error in estimates of the covariance matrix values between the three sets of polygons illustrated in Figure \ref{rectangle_example}. The estimated 95\% error is $\pm 0.0001$ due to uncertainty in true covariance. Isotropic EQ kernel, lengthscale = $1$km.}%, variance=$3$ (due to normalising).}
\label{rectangle_table}
\end{table}

We experimented briefly at higher dimensions, looking at the estimates of the covariance between a pair of 4-dimensional hyperspheres of \hl{radii one and two placed with centres three units apart,} so just touching. Using an isotropic EQ kernel (lengthscale=2.0) %, variance = 3.0) 
 we compared ten points to ten rectangles in each sphere and found that the estimated covariance using points (instead of hyper-rectangles) had roughly double the MAE (specifically the correct value was 314, with MAEs for points and rectangles were 50.0 and 25.1 respectively).

\section{Discussion}

In this paper we have derived both an analytical method for inference over cuboid integrals and an approximate method for inference over arbitrary inputs consisting of arbitrary sets of polytopes. In all the experiments, the integral kernels were able to improve on widely used alternatives. However, the improvement was most pronounced when the training data \hl{was binned into relatively few bins.} The first example, using age data from a census area, demonstrated most clearly why this method may perform more accurately than the `centroid' alternative; when the dataset has a peak or trough, the centroid method will fail to fully explain the bin integrals, and will have shallower responses to these changes than the data suggests is necessary. \hl{Using} the method to predict integrals (Section \ref{audience}) was particularly effective, when compared to the centroid alternative. One immediate use case would be estimating the number of young adults from the age histogram, for example, for making local-shop stocking decisions, etc; the centroid method would massively underestimate the number of people in their mid-20s. 

\hl{In some of the examples we model count data, this typically is non-negative, so we incorporate the work of} \citet{riihimaki2010gaussian} \hl{to enforce a non-negative latent function. This changes the posterior considerably and the ML estimates of the hyperparameters, thus influencing the entire domain. The practical utility of this operation probably depends on the dataset, for the example we used, the less-principled Gaussian-likelihood-only method performed slightly better.}

\hl{Other kernels could be substituted for the EQ. Although this requires some analytical integration work, we have found for other popular kernels the derivation straightforward. The supplementary contains an example of the exponential and linear kernel.}

Finally, in Section \ref{shapes}, we looked at approximation methods for non-cuboid, disjoint input regions. First we implemented the point-based approximation of \citet{kyriakidis2004geostatistical}. Although it did not achieve a particularly practical RMSE on the census dataset, it beat the centroid alternative, and provides a principled method for handling such data. However it is likely to be restricted to lower-dimensional spaces due to the increasing number of approximation points required in higher dimensions. We then replaced the approximation built of points with one built of rectangular patches, and used the covariance computed using the integral kernel. We found we needed considerably fewer rectangles than points to achieve similar accuracies. It is important to note though that the benefit from reduced numbers of training points is likely to be cancelled by the complexity of the integral kernel's covariance function, specifically the computation of four erfs in \eqref{kFF} and \eqref{kFf}. However the relative advantages depend on the shape being approximated. Clearly an L shape will probably be more efficiently approximated by two rectangles than by many randomly placed points. Further improvements are possible for more complex shapes, as we \hl{have not} used the most efficient rectangle placement algorithm. The rectangles could extend beyond the shape being approximated. One could introduce rectangles that contribute a negative weight, to delete those outlying regions, or cancel out patches where two rectangles have overlapped. We leave such enhancements for future researchers.

%A problem optimising hyperparameters exists with the current implementation of the algorithm in section \ref{shapes}. Experimentally we found that the noise inherent in the random placement of points caused the gradient optimisation of the hyperparameters to become quite problematic, and caused issues with standard stopping criteria. We were placing new points each optimisation step however, thus an obvious future mitigation would be to ensure the same integration points are chosen throughout the optimisation phase.

In this paper we have proposed and derived principled and effective methods for analytical and approximate inference over binned datasets. We have tested these methods on several datasets and found them to be effective and superior to alternatives. This provides an easy, useful and principled toolkit for researchers and developers handling histogrammed or binned datasets, who wish to improve their prediction accuracies.

%\begin{acknowledgements}
%This work has been supported by the Engineering and Physical Research Council (EPSRC) Research Project EP/N014162/1. We also thank Wil Ward for assistance in the writing of this paper.
%\end{acknowledgements}

\bibliographystyle{spbasic}      % basic style, author-year citations

\bibliography{refs}

\begin{thebibliography}{20}
\providecommand{\natexlab}[1]{#1}
\providecommand{\url}[1]{{#1}}
\providecommand{\urlprefix}{URL }
\expandafter\ifx\csname urlstyle\endcsname\relax
  \providecommand{\doi}[1]{DOI~\discretionary{}{}{}#1}\else
  \providecommand{\doi}{DOI~\discretionary{}{}{}\begingroup
  \urlstyle{rm}\Url}\fi
\providecommand{\eprint}[2][]{\url{#2}}

\bibitem[{Alt et~al(1995)Alt, Hsu, and Snoeyink}]{alt1995computing}
Alt H, Hsu D, Snoeyink J (1995) Computing the largest inscribed isothetic
  rectangle. In: Canadian Conference on Computational Geometry, pp 67--72

\bibitem[{Alvarez et~al(2009)Alvarez, Luengo, and Lawrence}]{alvarez2009latent}
Alvarez M, Luengo D, Lawrence N (2009) Latent force models. In: Artificial
  Intelligence and Statistics, pp 9--16

\bibitem[{A{\v{z}}man and Kocijan(2005)}]{avzman2005comprising}
A{\v{z}}man K, Kocijan J (2005) Comprising prior knowledge in dynamic
  {G}aussian process models. In: Proceedings of the International Conference on
  Computer Systems and Technologies, vol~16

\bibitem[{Beranger et~al(2018)Beranger, Lin, and Sisson}]{beranger2018new}
Beranger B, Lin H, Sisson SA (2018) New models for symbolic data analysis.
  arXiv preprint arXiv:180903659

\bibitem[{Cabello et~al(2016)Cabello, Cheong, Knauer, and
  Schlipf}]{cabello2016finding}
Cabello S, Cheong O, Knauer C, Schlipf L (2016) Finding largest rectangles in
  convex polygons. Computational Geometry 51:67--74

\bibitem[{Calder and Cressie(2007)}]{calder2007some}
Calder CA, Cressie N (2007) Some topics in convolution-based spatial modeling.
  Proceedings of the 56th Session of the International Statistics Institute pp
  22--29

\bibitem[{Daniels et~al(1997)Daniels, Milenkovic, and
  Roth}]{daniels1997finding}
Daniels KL, Milenkovic VJ, Roth D (1997) Finding the largest area axis-parallel
  rectangle in a polygon. Computational Geometry 7:125--148

\bibitem[{Dwork and Roth(2014)}]{dwork2014algorithmic}
Dwork C, Roth A (2014) The algorithmic foundations of differential privacy.
  Foundations and Trends in Theoretical Computer Science 9(3-4):211--407

\bibitem[{Gosling et~al(2007)Gosling, Oakley, {O'Hagan}
  et~al}]{gosling2007nonparametric}
Gosling JP, Oakley JE, {O'Hagan} A, et~al (2007) Nonparametric elicitation for
  heavy-tailed prior distributions. Bayesian Analysis 2(4):693--718

\bibitem[{Grimme(2015)}]{grimme2015picking}
Grimme C (2015) Picking a uniformly random point from an arbitrary simplex.
  \url{https://www.researchgate.net/profile/Christian_Grimme/publication/275348534_Picking_a_Uniformly_Random_Point_from_an_Arbitrary_Simplex/links/553a08800cf247b858815a6b.pdf},
  {U}niversity of M{\"u}nster [Online; accessed 31-July-2018]

\bibitem[{Iacob et~al(2003)Iacob, Marinescu, and Luca}]{iacob2003covering}
Iacob P, Marinescu D, Luca C (2003) Covering with rectangular pieces. Analele
  Stiintifice ale Universitatii Ovidius Constanta 11(2):75--86

\bibitem[{Knauer et~al(2012)Knauer, Schlipf, Schmidt, and
  Tiwary}]{knauer2012largest}
Knauer C, Schlipf L, Schmidt JM, Tiwary HR (2012) Largest inscribed rectangles
  in convex polygons. Journal of discrete algorithms 13:78--85

\bibitem[{Kyriakidis(2004)}]{kyriakidis2004geostatistical}
Kyriakidis PC (2004) A geostatistical framework for area-to-point spatial
  interpolation. Geographical Analysis 36(3):259--289

\bibitem[{Le-Rademacher and Billard(2017)}]{le2017principal}
Le-Rademacher J, Billard L (2017) Principal component analysis for
  histogram-valued data. Advances in Data Analysis and Classification
  11(2):327--351

\bibitem[{Oakley and {O'Hagan}(2007)}]{oakley2007uncertainty}
Oakley JE, {O'Hagan} A (2007) Uncertainty in prior elicitations: a
  nonparametric approach. Biometrika 94(2):427--441

\bibitem[{{O'Hagan}(1991)}]{o1991bayes}
{O'Hagan} A (1991) Bayes--{H}ermite quadrature. Journal of statistical planning
  and inference 29(3):245--260

\bibitem[{Ramsay(2006)}]{ramsay2006functional}
Ramsay JO (2006) Functional data analysis, Chapter 16. Wiley Online Library

\bibitem[{Riihim{\"a}ki and Vehtari(2010)}]{riihimaki2010gaussian}
Riihim{\"a}ki J, Vehtari A (2010) Gaussian processes with monotonicity
  information. In: Proceedings of the Thirteenth International Conference on
  Artificial Intelligence and Statistics, pp 645--652

\bibitem[{Stein(1966)}]{stein1966note}
Stein P (1966) A note on the volume of a simplex. The American Mathematical
  Monthly 73(3):299--301

\bibitem[{Williams and Rasmussen(2006)}]{williams2006gaussian}
Williams CK, Rasmussen CE (2006) Gaussian processes for machine learning. the
  MIT Press

\end{thebibliography}
\clearpage

\end{document}